\documentclass[11pt,a4paper]{article}
\usepackage[hyperref]{acl2020}
\usepackage{times}
\usepackage{latexsym}
\usepackage{graphicx} % DO NOT CHANGE THIS
\usepackage{color}
\usepackage{amssymb}
\usepackage{amsmath}
\usepackage{algorithm}
\usepackage{algorithmic}
\usepackage{subfigure}
\usepackage{multirow}
\usepackage{booktabs}
\usepackage{xcolor}%定义了一些颜色  
\usepackage{colortbl,booktabs}%第二个包定义了几个*rule  
\usepackage{microtype}

\aclfinalcopy % Uncomment this line for the final submission
\def\aclpaperid{120} %  Enter the acl Paper ID here

\title{Relational Graph Attention Network for Aspect-based Sentiment Analysis}

\author{Kai Wang\textsuperscript{1}, Weizhou Shen\textsuperscript{1}, Yunyi Yang\textsuperscript{1}, Xiaojun Quan\textsuperscript{1}\thanks{\; Corresponding author.}, Rui Wang\textsuperscript{2}\\
\textsuperscript{1}School of Data and Computer Science, Sun Yat-sen University, China \\
\textsuperscript{2}Alibaba Group, China\\
\tt \{wangk73,~shenwzh3,~yangyy37\}@mail2.sysu.edu.cn \\
\tt quanxj3@mail.sysu.edu.cn, \tt masi.wr@alibaba-inc.com}

\date{}

\begin{document}
\maketitle
\begin{abstract}
Aspect-based sentiment analysis aims to determine the sentiment polarity towards a specific aspect in online reviews.~Most recent efforts adopt attention-based neural network models to implicitly connect aspects with opinion words. However, due to the complexity of language and the existence of multiple aspects in a single sentence, these models often confuse the connections. In this paper, we address this problem by means of effective encoding of syntax information. Firstly, we define a unified aspect-oriented dependency tree structure rooted at a target aspect by reshaping and pruning an ordinary dependency parse tree. Then, we propose a relational graph attention network (R-GAT) to encode the new tree structure for sentiment prediction.~Extensive experiments are conducted on the SemEval 2014 and Twitter datasets, and the experimental results confirm that the connections between aspects and opinion words can be better established with our approach, and the performance of the graph attention network (GAT) is significantly improved as a consequence.
\end{abstract}

\section{Introduction}
Aspect-based sentiment analysis (ABSA) aims at fine-grained sentiment analysis of online affective texts such as product reviews.~Specifically, its objective is to determine the sentiment polarities towards one or more aspects appearing in a single sentence.~An example of this task is, given a review \texttt{great food but the service was dreadful},~to determine the polarities towards the aspects \texttt{food} and \texttt{service}. Since the two aspects express quite opposite sentiments, just assigning a sentence-level sentiment polarity is inappropriate. In this regard, ABSA can provide better insights into user reviews compared with sentence-level sentiment analysis.

Intuitively, connecting aspects with their respective opinion words lies at the heart of this task. Most recent efforts  \cite{wang2016attention,li2017deep,ma2017interactive,fan2018multi} resort to assorted attention mechanisms to achieve this goal and have reported appealing results. However, due to the complexity of language morphology and syntax, these mechanisms fail occasionally. We illustrate this problem with a real review \texttt{So delicious was the noodles but terrible vegetables}, in which the opinion word \texttt{terrible} is closer to the aspect \texttt{noodles} than \texttt{delicious}, and there could be \texttt{terrible noodles} appearing in some other reviews which makes these two words closely associated. Therefore, the attention mechanisms could attend to \texttt{terrible} with a high weight when evaluating the aspect \texttt{noodles}.

Some other efforts explicitly leverage the syntactic structure of a sentence to establish the connections. Among them, early attempts rely on hand-crafted syntactic rules \cite{qiu2011opinion,liu2013opinion}, though they are subject to the quantity and quality of the rules. Dependency-based parse trees are then used to provide more comprehensive syntactic information. For this purpose, a whole dependency tree can be encoded from leaves to root by a recursive neural network (RNN) \cite{lakkaraju2014aspect,dong2014adaptive,nguyen2015phrasernn,wang2016recursive}, or the internal node distance can be computed and used for attention weight decay \cite{he2018effective}. Recently, graph neural networks (GNNs) are explored to learn representations from the dependency trees \cite{zhang-etal-2019-aspect,sun-etal-2019-aspect,huang-carley-2019-syntax}. The shortcomings of these approaches should not be overlooked. First, the dependency
relations, which may indicate the connections between aspects and opinion words, are ignored. Second, empirically, only a small part of the parse tree is related to this task and it is unnecessary to encode the whole tree \cite{Zhang0M18,ZhaoHLB18}. Finally, the encoding process is tree-dependent, making the batch operation inconvenient during optimization. 

In this paper, we re-examine the syntax information and claim that revealing task-related syntactic structures is the key to address the above issues. We propose a novel aspect-oriented dependency tree structure constructed in three steps.~Firstly, we obtain the dependency tree of a sentence using an ordinary parser. Secondly, we reshape the dependency tree to root it at a target aspect in question. Lastly, pruning of the tree is performed to retain only edges with direct dependency relations with the aspect. Such a unified tree structure not only enables us to focus on the connections between aspects and potential opinion words but also facilitates both batch and parallel operations. Then we propose a relational graph attention network (R-GAT) model to encode the new dependency trees. R-GAT generalizes graph attention network (GAT) to encode graphs with labeled edges. Extensive evaluations are conducted on the SemEval 2014 and Twitter datasets, and experimental results show that R-GAT significantly improves the performance of GAT. It also achieves superior performance to the baseline methods. 

The contributions of this work include:
\vspace{-0.14cm}
\begin{itemize}
\setlength\itemsep{-0.05cm}
  \item We propose an aspect-oriented tree structure by reshaping and pruning ordinary dependency trees to focus on the target aspects.
 \item We propose a new GAT model to encode the dependency relations and to establish the connections between aspects and opinion words.
  \item The source code of this work is released for future research.\footnote{https://github.com/shenwzh3/RGAT-ABSA}
\end{itemize}

\section{Related Work}
Most recent research work on aspect-based sentiment analysis (ABSA) utilizes attention-based neural models to examine words surrounding a target aspect.~They can be considered an implicit approach to exploiting sentence structure, since opinion words usually appear not far from aspects. Such approaches have led to promising progress. Among them, Wang et al. \shortcite{wang2016attention} proposed to use an attention-based LSTM to identify important sentiment information relating to a target aspect. Chen et al. \shortcite{chen2017recurrent} introduced a multi-layer attention mechanism to capture long-distance opinion words for aspects. For a similar purpose, Tang et al. \shortcite{tang2016aspect} employed Memory Network with multi-hop attention and external memory. Fan et al. \shortcite{fan2018multi} proposed a multi-grained attention network with both fine-grained and coarse-grained attentions. The pre-trained language model BERT \cite{devlin2018bert} has made successes in many classification tasks including ABSA. For example, \newcite{xu2019bert} used an additional corpus to post-train BERT and proved its effectiveness in both aspect extraction and ABSA. \newcite{sun2019utilizing} converted ABSA to a sentence-pair classification task by constructing auxiliary sentences.

Some other efforts try to directly include the syntactic information in ABSA. Since aspects are generally assumed to lie at the heart of this task, establishing the syntactic connections between each target aspect and the other words are crucial. Qiu et al. \shortcite{qiu2011opinion} manually defined some syntactic rules to identify the relations between aspects and potential opinion words. Liu et al. \shortcite{liu2013opinion} obtained partial alignment links with these syntactic rules and proposed a partially supervised word alignment model to extract opinion targets. Afterward, neural network models were explored for this task. Lakkaraju et al. \shortcite{lakkaraju2014aspect} used a recursive neural network (RNN) to hierarchically encode word representations and to jointly extract aspects and sentiments. In another work, Wang et al. \shortcite{wang2016recursive} combined the recursive neural network with conditional random fields (CRF). Moreover, Dong et al. \shortcite{dong2014adaptive} proposed an adaptive recursive neural network (AdaRNN) to adaptively propagate the sentiments of words to the target aspect via semantic composition over a dependency tree. Nguyen et al. \shortcite{nguyen2015phrasernn} further combined the dependency and constituent trees of a sentence with a phrase recursive neural network (PhraseRNN). In a simpler approach, He et al. \shortcite{he2018effective} used the relative distance in a dependency tree for attention weight decay. They also showed that selectively focusing on a small subset of context words can lead to satisfactory results.

\begin{figure*}[!ht]
\centering
\subfigure[]{
\label{fig:1a}
\includegraphics[height=2cm]{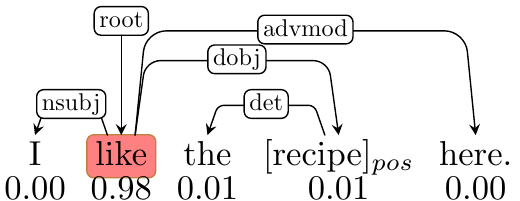}
}
\subfigure[]{
 %% label for second subfigure
 \label{fig:1b}
\includegraphics[height=2cm]{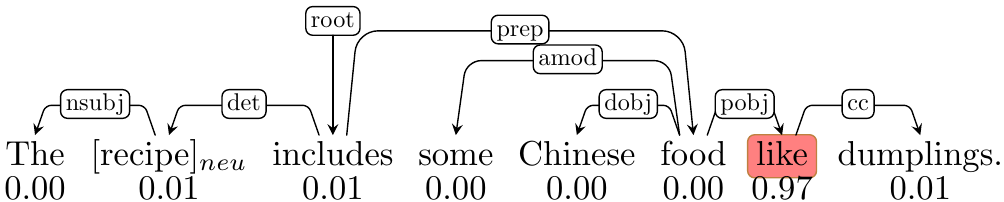}
}
\subfigure[]{
\label{fig:1c}
\includegraphics[width=13cm]{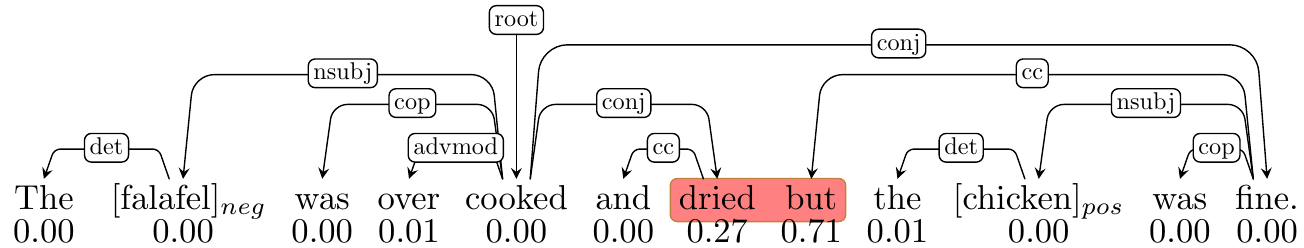}
}
\caption{Three examples from restaurant reviews to illustrate the relationships among aspect, attention, and syntax in ABSA. Labeled edges indicate dependency relations, and scores under each word represent attention weights assigned by the attention-equipped LSTM. Words with high attention weights are highlighted in red boxes, and words in brackets are the target aspects followed by their sentiment labels.}
\label{img:dependency example}
\end{figure*}

Recently, graph neural networks combined with dependency trees have shown appealing effectiveness in ABSA. \newcite{zhang-etal-2019-aspect} and \newcite{sun-etal-2019-aspect} proposed to use graph convolutional networks (GCN) to learn node representations from a dependency tree and used them together with other features for sentiment classification. For a similar purpose, \newcite{huang-carley-2019-syntax} used graph attention networks (GAT) to explicitly establish the dependency relationships between words. However, these approaches generally ignore the dependency relations which might identify the connections between aspects and opinion words.

\section{Aspect-Oriented Dependency Tree}
In this section, we elaborate on the details of constructing an aspect-oriented dependency tree.

\subsection{Aspect, Attention and Syntax}\label{sec:example}
The syntactic structure of a sentence can be uncovered by dependency parsing, a task to generate a dependency tree to represent the grammatical structure. The relationships between words can be denoted with directed edges and labels. We use three examples to illustrate the relationships among aspect, attention and syntax in ABSA, as shown in Figure \ref{img:dependency example}. In the first example, the word \texttt{like} is used as a verb and it expresses a positive sentiment towards the aspect \texttt{recipe}, which is successfully attended by the attention-based LSTM model. However, when it is used as a preposition in the second example, the model still attends to it with a high weight, resulting in a wrong prediction. The third example shows a case where there are two aspects in a single sentence with different sentiment polarities. For the aspect \texttt{chicken}, the LSTM model mistakenly assigns high attention weights to the words \texttt{but} and \texttt{dried}, which leads to another prediction mistake. These examples demonstrate the limitations of the attention-based model in this task. Such mistakes are likely to be avoided by introducing explicit syntactic relations between aspects and other words. For example, it might be different if the model noticed the direct dependency relationship between \texttt{chicken} and \texttt{fine} in the third example, rather than with \texttt{but}.

\subsection{Aspect-Oriented Dependency Tree}
The above analysis suggests that dependency relations with direct connections to an aspect may assist a model to focus more on related opinion words, and therefore should be more important than other relations. Also, as shown in Figure \ref{img:dependency example}, a dependency tree contains abundant grammar information, and is usually not rooted at a target aspect. Nevertheless, the focus of ABSA is a target aspect rather than the root of the tree.  Motivated by the above observations, we propose a novel aspect-oriented dependency tree structure by reshaping an original dependency tree to root it at a target aspect, followed by pruning of the tree so as to discard unnecessary relations.

\begin{figure*}[t]
\centering
\includegraphics[width=13.5cm]{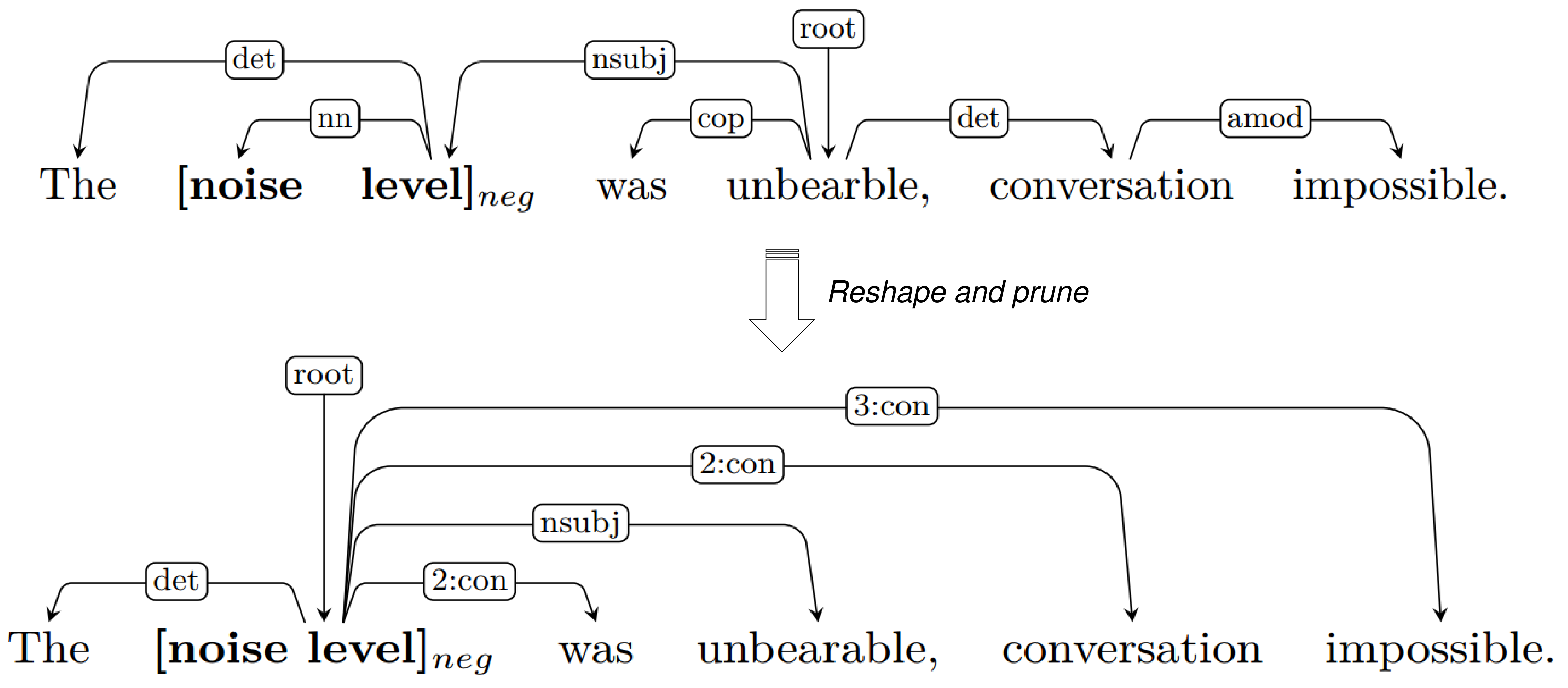}
\caption{Construction of an aspect-oriented dependency tree (bottom) from an ordinary dependency tree (top). 
}
\label{img:reshape}
\end{figure*}

\begin{algorithm}[h]
\caption{Aspect-Oriented Dependency Tree}
\label{algorithm:Construct the Unified Tree}
\begin{algorithmic}[1] %这个1 表示每一行都显示数字
\REQUIRE aspect $a=\{w_i^a, w_{i+1}^a, ... w_k^a\}$, sentence $s=\{w_1^s, w_2^s, ... w_n^s\}$, dependency tree $T$, and  dependency relations $r$.
\ENSURE aspect-oriented dependency tree $\hat{T}$.
\STATE Construct the root $R$ for $\hat{T}$;
\FOR{$i$ to $k$}
\FOR{$j=1$ to $n$}
\IF{$w_j^s\xrightarrow{r_{ji}} w_i^a$}
\STATE $w_j^s\xrightarrow{r_{ji}} R$
\ELSIF{$w_j^s\xleftarrow{r_{ij}} w_i^a$}
\STATE $w_j^s\xleftarrow{r_{ij}} R$
\ELSE
\STATE $n=distance(i,j)$
\STATE $w_j^s\xrightarrow{n:con} R$
\ENDIF
\ENDFOR
\ENDFOR
\RETURN $\hat{T}$ %算法的返回值
\end{algorithmic}
\end{algorithm}

Algorithm \ref{algorithm:Construct the Unified Tree} describes the above process. For an input sentence, we first apply a dependency parser to obtain its dependency tree, where $r_{ij}$ is the dependency relation from node $i$ to $j$. Then, we build an aspect-oriented dependency tree in three steps. Firstly, we place the target aspect at the root, where multiple-word aspects are treated as entities. Secondly, we set the nodes with direct connections to the aspect as the children, for which the original dependency relations are retained. Thirdly, other dependency relations are discarded, and instead, we put a virtual relation \texttt{n:con} ($n$ connected) from the aspect to each corresponding node, where $n$ represents the distance between two nodes.\footnote{We set $n=\infty$ if the distance is longer than 4.} If the sentence contains more than one aspect, we construct a unique tree for each aspect. Figure \ref{img:reshape} shows an aspect-oriented dependency tree constructed from the ordinary dependency tree. There are at least two advantages with such an aspect-oriented structure. First, each aspect has its own dependency tree and can be less influenced by unrelated nodes and relations. Second, if an aspect contains more than one word, the dependency relations will be aggregated at the aspect, unlike in \cite{zhang-etal-2019-aspect,sun-etal-2019-aspect} which require extra pooling or attention operations.

%We further take the sentence in Figure \ref{fig:1c} as another example to illustrate the above algorithm and process. Suppose there are only three dependency relations: \texttt{advmod}, \texttt{nsubj} and \texttt{cop}. The aspect \texttt{falafel} in this example has two directly connected nodes, \texttt{The} and \texttt{cooked}, in the original dependency tree. To reshape this tree, we first make the aspect \texttt{falafel} a new root. For the node \texttt{cooked}, we retain its relation with the root \texttt{falafel} in the new tree. Since we define a new relation \texttt{ncon}, the relation set now includes \texttt{advmod}, \texttt{nsubj}, \texttt{cop} and \texttt{ncon}. If the original probabilities over the three relations from \texttt{cooked} to \texttt{falafel} are $[0.1, 0.8, 0.1]$, the probabilities will change to $[0.1, 0.8, 0.1, 0]$ after the reshaping operation, with \texttt{ncon} having a zero probability. For those unconnected nodes with \texttt{falafel} like \texttt{was} and \texttt{over}, we connect them to \texttt{falafel} under the relation \texttt{ncon} with probabilities $[0, 0, 0, 1]$.

The idea described above is partially inspired by previous findings \cite{he2018effective,Zhang0M18,ZhaoHLB18} that it could be sufficient to focus on a small subset of context words syntactically close to the target aspect. Our approach provides a direct way to model the context information. Such a unified tree structure not only enables our model to focus on the connections between aspects and opinion words but also facilitates both batch and parallel operations during training. The motivation we put a new relation \texttt{n:con} is that existing parsers may not always parse sentences correctly and may miss important connections to the target aspect. In this situation, the relation \texttt{n:con} enables the new tree to be more robust. We evaluate this new relation in the experiment and the results confirm this assumption.

\section{Relational Graph Attention Network}
To encode the new dependency trees for sentiment analysis, we propose a relational graph attention network (R-GAT) by extending the graph attention network (GAT) \cite{velivckovic2017graph} to encode graphs with labeled edges. 

\subsection{Graph Attention Network}
Dependency tree can be represented by a graph $\mathcal{G}$ with $n$ nodes, where each represents a word in the sentence. The edges of $\mathcal{G}$ denote the dependency between words. The neighborhood nodes of node $i$ can be represented by $\mathcal{N}_i$. GAT iteratively updates each node representation (e.g., word embeddings) by aggregating neighborhood node representations using multi-head attention:
\begin{equation}
     h_{att_i}^{l+1}=||_{k=1}^{K}\sum\limits_{j\in \mathcal{N}_i}\alpha_{ij}^{lk}W_k^lh_j^l
\end{equation}
 \begin{equation}
  \alpha_{ij}^{lk}=attention(i,j)
  \end{equation}
where $h_{att_i}^{l+1}$ is the attention head of node $i$ at layer $l+1$, $||_{k=1}^{K}x_i$ denotes the concatenation of vectors from $x_1$ to $x_k$, $\alpha_{ij}^{lk}$ is a normalized attention coefficient computed by the $k$-th attention at layer $l$, $W_k^l$ is an input transformation matrix. In this paper, we adopt dot-product attention for $attention(i,j)$.\footnote{Dot product has fewer parameters but similar performance with feedforward neural network used in \cite{velivckovic2017graph}.}

\subsection{Relational Graph Attention Network}
\begin{figure}[h]
\centering
\includegraphics[scale=0.5]{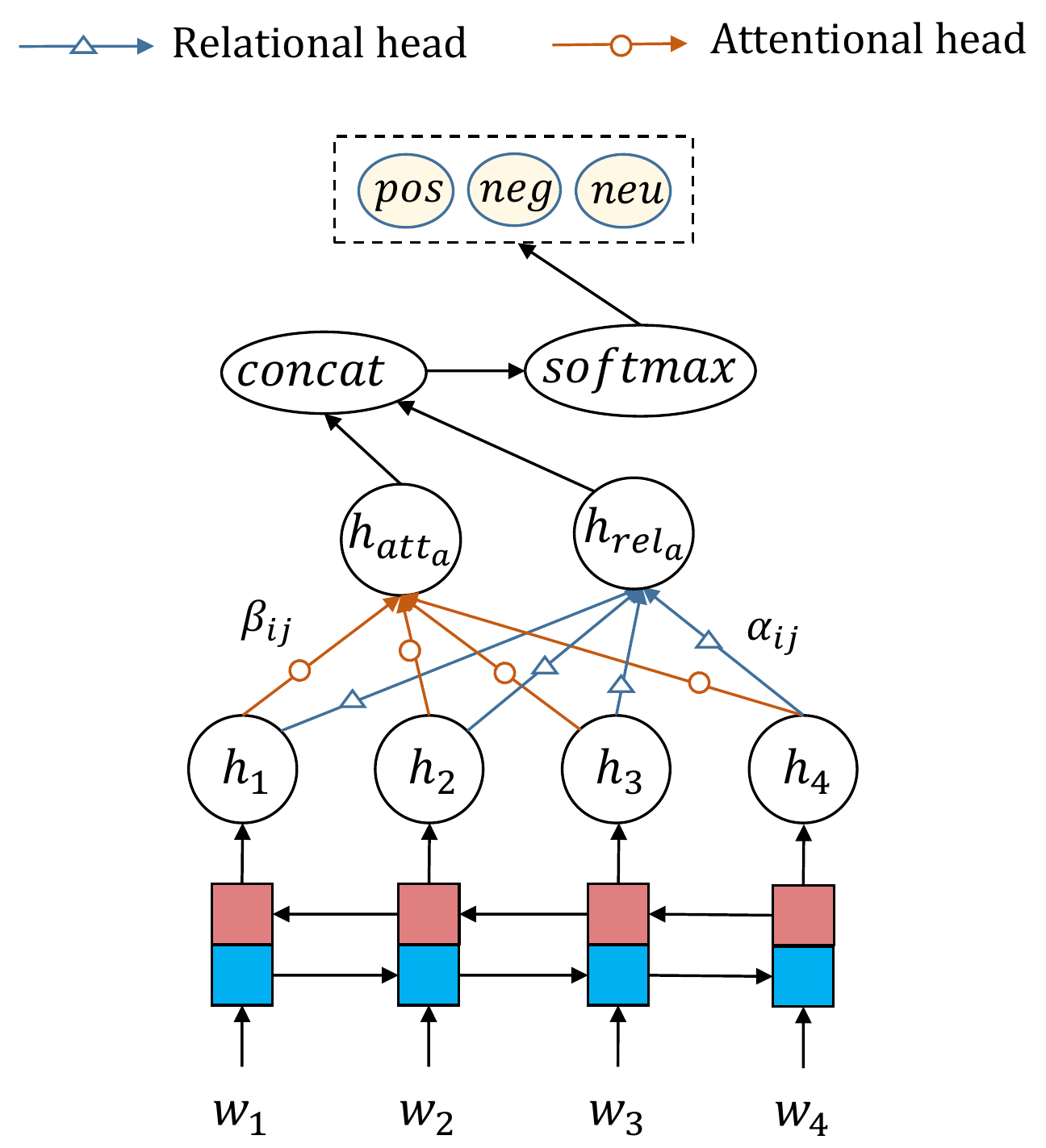}
\caption{Structure of the proposed relational graph attention network (R-GAT), which includes two genres of multi-head attention mechanism, i.e., attentional head and relational head. }
\label{img:model}
\end{figure}

GAT aggregates the representations of neighborhood nodes along the dependency paths. However, this process fails to take dependency relations into consideration, which may lose some important dependency information. Intuitively, neighborhood nodes with different dependency relations should have different influences. We propose to extend the original GAT with additional relational heads. We use these relational heads as relation-wise gates to control information flow from neighborhood nodes. The overall architecture of this approach is shown in Figure \ref{img:model}. Specifically, we first map the dependency relations into vector representations, and then compute a relational head as:
\begin{gather}
  h_{rel_i}^{l+1}=||_{m=1}^{M}\sum\limits_{j\in \mathcal{N}_i}\beta_{ij}^{lm}W_m^lh_j^l\\
  g_{ij}^{lm}=\sigma(relu(r_{ij}W_{m1}+b_{m1})W_{m2}+b_{m2})\\
  \beta_{ij}^{lm}=\frac{exp(g_{ij}^{lm})}{\sum_{j=1}^{\mathcal{N}_i}exp(g_{ij}^{lm})}
\end{gather}
where $r_{ij}$ represents the relation embedding between nodes $i$ and $j$. R-GAT contains $K$ attentional heads and $M$ relational heads. The final representation of each node is computed by:
\begin{gather}
 x_i^{l+1}=h_{att_i}^{l+1}\ ||\ h_{rel_i}^{l+1}\\
 h_i^{l+1}=relu(W_{l+1}x_i^{l+1}+b_{l+1})
\end{gather}

\subsection{Model Training}
We use BiLSTM to encode the word embeddings of tree nodes, and obtain its output hidden state $h_i$ for the initial representation $h^0_i$ of leaf node $i$. Then, another BiLSTM is applied to encode the aspect words, and its average hidden state is used as the initial representation $h^0_a$ of this root. After applying R-GAT on an aspect-oriented tree, its root representation $h_a^{l}$ is passed through a fully connected softmax layer and mapped to probabilities over the different
sentiment polarities.
\begin{equation}
  p(a)=softmax(W_ph_a^l+b_p)
\end{equation}
Finally, the standard cross-entropy loss is used as our objective function:
\begin{equation}
  L(\theta)=-\sum\limits_{(S,A)\in \mathcal{D}}\sum\limits_{a\in A}\log p(a)
\end{equation}
where $\mathcal{D}$ contains all the sentence-aspects pairs, $A$ represents the aspects appearing in sentence $S$, and $\theta$ contains all the
trainable parameters.

\section{Experiments}
In this section, we first introduce the datasets used for evaluation and the baseline methods employed for comparison. Then, we report the experimental results conducted from different perspectives. Finally, error analysis and discussion are conducted with a few representative examples.

\subsection{Datasets}
Three public sentiment analysis datasets are used in our experiments, two of them are the \texttt{Laptop} and \texttt{Restaurant} review datasets from the SemEval 2014 Task \cite{pontiki2014semeval},\footnote{http://alt.qcri.org/semeval2014/task4/.} and the third is the \texttt{Twitter} dataset used by \cite{dong2014adaptive}. Statistics of the three datasets can be found in Table \ref{tab:dataset}.

\begin{table}[t]
\centering
\resizebox{1\columnwidth}{!}{%
\begin{tabular}{lccccccccc}
\toprule
\multirow{2}{*}{Dataset} & \multicolumn{2}{c}{Positive} & \multicolumn{2}{c}{Neutral} & \multicolumn{2}{c}{Negative} \\
\cmidrule(r){2-3} \cmidrule(r){4-5} \cmidrule(r){6-7}
&  Train      &  Test
&  Train      &  Test
&  Train      &  Test   \\
\midrule
Laptop &994&341&870&128&464&169\\
Restaurant &2164&728&807&196&637&196\\
Twitter &1561&173&3127&346&1560&173\\
\bottomrule
\end{tabular}
}
\caption{Statistics of the three datasets.}\vspace{-0.5cm}
\label{tab:dataset}
\end{table}

\subsubsection{Implementation Details}
 The Biaffine Parser \cite{dozat2016deep} is used for dependency parsing. The dimension of the dependency relation embeddings is set to 300. For R-GAT, we use the 300-dimensional word embeddings of GLoVe \cite{pennington2014glove}. For R-GAT+BERT, we use the last hidden states of the pre-trained BERT for word representations and fine-tune them on our task. The PyTorch implementation of BERT \footnote{https://github.com/huggingface/transformers} is used in the experiments. R-GAT is shown to prefer a high dropout rate in between [0.6, 0.8]. As for R-GAT+BERT, it works better with a low dropout rate of around 0.2. Our model is trained using the Adam optimizer \cite{kingma2014adam} with the default configuration.  
 
\subsection{Baseline Methods}
A few mainstream models for aspect-based sentiment analysis are used for comparison, including:

\begin{itemize}
\setlength\itemsep{-0.04cm}
\item \textbf{Syntax-aware models:} LSTM+SynATT \cite{he2018effective}, AdaRNN \cite{dong2014adaptive}, PhraseRNN \cite{nguyen2015phrasernn}, ASGCN \cite{zhang-etal-2019-aspect}, CDT \cite{sun-etal-2019-aspect}, GAT \cite{velivckovic2017graph} and TD-GAT \cite{huang-carley-2019-syntax}.
\item \textbf{Attention-based~models:}~ATAE-LSTM \cite{wang2016attention} , {IAN} \cite{ma2017interactive}, {RAM} \cite{chen2017recurrent}, {MGAN} \cite{fan2018multi}, attention-equipped {LSTM}, and fine-tuned {BERT} \cite{devlin2018bert}.
\item \textbf{Other recent methods:} {GCAE} \cite{xue2018aspect}, {JCI} \cite{wang2018target} and TNET \cite{li2018transformation}.
\item \textbf{Our methods:} R-GAT is our relational graph attention network.~R-GAT+BERT is our R-GAT with the BiLSTM replaced by BERT, and the attentional heads of R-GAT will also be replaced by that of BERT.
\end{itemize}

\begin{table*}[h]
\centering
\resizebox{1.9\columnwidth}{!}{%
\begin{tabular}{clcccccc}
\toprule
\multirow{2}{*}{Category} &\multirow{2}{*}{Method} & \multicolumn{2}{c}{Restaurant} & \multicolumn{2}{c}{Laptop} & \multicolumn{2}{c}{Twitter} \\
\cmidrule(r){3-4} \cmidrule(r){5-6} \cmidrule(r){7-8}
& &  Accuracy      &  Macro-F1&  Accuracy      &  Macro-F1 &  Accuracy      &  Macro-F1   \\
\midrule
\multirow{7}{*}{Syn.}&LSTM+SynATT  &80.45&71.26&72.57&69.13&-&-\\
&AdaRNN  &-&-&-&-& 66.30&65.90\\
 &PhraseRNN  &66.20&59.32&-&-&-&-\\
 &ASGCN&80.77&72.02&75.55&71.05&72.15&70.40\\
 &CDT&82.30&74.02&77.19&72.99&74.66&73.66\\
  &GAT&78.21&67.17&73.04&68.11&71.67&70.13\\
  &TD-GAT&80.35&76.13&74.13&72.01&72.68&71.15\\
 \midrule
\multirow{6}{*}{Att.}&ATAE-LSTM  &77.20&-&68.70&-&-&-\\
&IAN  &78.60&-&72.10&-&-&-\\
&RAM  &80.23&70.80&74.49&71.35&69.36&67.30\\
&MGAN  &81.25&71.94&75.39&72.47&72.54&70.81\\
&LSTM  &79.10&69.00&71.22&65.75&69.51&67.98\\
&BERT &85.62&78.28&77.58&72.38&75.28&74.11\\
\midrule
\multirow{3}{*}{Others}&GCAE &77.28&-&69.14&-&-&-\\
&JCI &-&68.84&-&67.23&-&-\\
&TNET  &80.69&71.27&76.54&71.75&74.90&73.60\\
\midrule
\rowcolor{gray!10} \multirow{1}{*}{Ours}
 &R-GAT &83.30&76.08&77.42&73.76&75.57&73.82\\
\rowcolor{gray!10} \multirow{1}{*}{Ours}
 &R-GAT+BERT &\textbf{86.60}&\textbf{81.35}&\textbf{78.21}&\textbf{74.07}&\textbf{76.15}&\textbf{74.88}\\ 
\bottomrule
\end{tabular}
}
\caption{Overall performance of different methods on the three datasets.}
\label{tab:overall}
\end{table*}

\subsection{Results and Analysis}

\subsubsection{Overall Performance}
The overall performance of all the models are shown in Table \ref{tab:overall}, from which several observations can be noted. First, the R-GAT model outperforms most of the baseline models. Second, the performance of GAT can be significantly improved when incorporated with relational heads in our aspect-oriented dependency tree structure. It also outperforms the baseline models of ASGCN, and CDT, which also involve syntactic information in different ways. This proves that our R-GAT is better at encoding the syntactic information. Third, the basic BERT can already outperform all the existing ABSA models by significant margins, demonstrating the power of this large pre-trained model in this task. Nevertheless, after incorporating our R-GAT (R-GAT+BERT), this strong model sees further improvement and has achieved a new state of the art. These results have demonstrated the effectiveness of our R-GAT in capturing important syntactic structures for sentiment analysis.

\subsubsection{Effect of Multiple Aspects}

\begin{figure}[!ht]
\centering
\includegraphics[width=7.7cm]{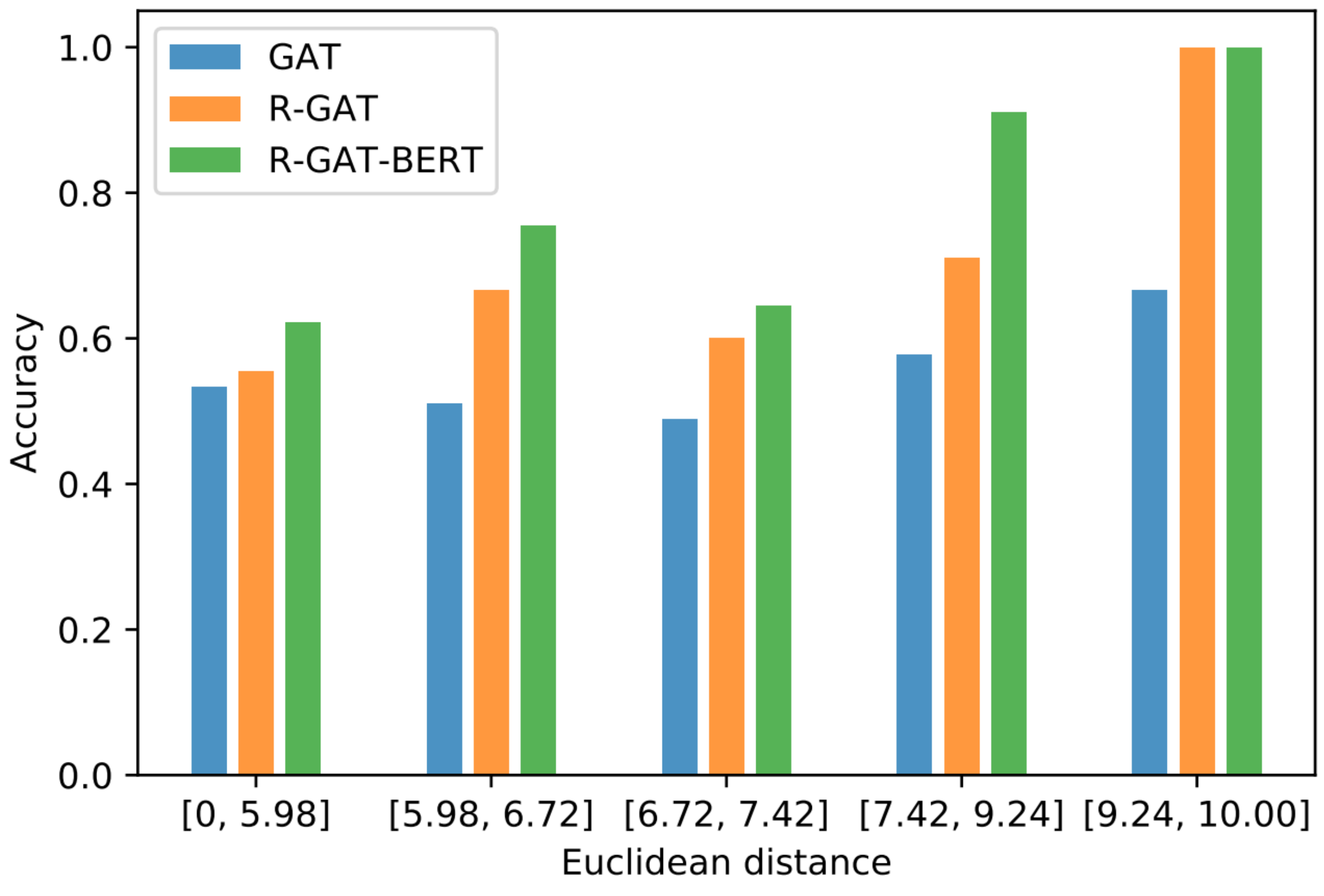}
\caption{Results of multiple aspects analysis, which shows that the aspects with nearer distances tend to lead to lower accuracy scores. }
\label{experiment:multiple aspects}
\end{figure}

The appearance of multiple aspects in one single sentence is very typical for ABSA. To study the influence of multiple aspects, we pick out the reviews with more than one aspect in a sentence. Each aspect is represented with its averaged (GloVe) word embeddings, and the distance between any two aspects of a sentence is calculated using the Euclidean distance. If there are more than two aspects, the nearest Euclidean distance is used for each aspect. Then, we select three models (GAT, R-GAT, R-GAT+BERT) for sentiment prediction, and plot the aspect accuracy by different distance ranges in Figure \ref{experiment:multiple aspects}. We can observe that the aspects with nearer distances tend to lead to lower accuracy scores, indicating that the aspects with high semantic similarity in a sentence may confuse the models. However, with our R-GAT, both GAT and BERT can be improved across different ranges, showing that our method can alleviate this problem to a certain extent.

\subsubsection{Effect of Different Parsers}

Dependency parsing plays a critical role in our method. To evaluate the impact of different parsers, we conduct a study based on the R-GAT model using two well-known dependency parsers: Stanford Parser \cite{chen2014fast} and Biaffine Parser \cite{dozat2016deep}.\footnote{The parsers are implemented by Stanford CoreNLP \cite{manning2014stanford} and AllenNLP \cite{gardner2018allennlp}.} Table \ref{tab:parser} shows the performance of the two parsers in UAS and LAS metrics, followed by their performance for aspect-based sentiment analysis. From the table, we can find that the better Biaffine parser results in higher sentiment classification accuracies. Moreover, it further implies that while existing parsers can capture most of the syntactic structures correctly, our method has the potential to be further improved with the advances of parsing techniques.

\begin{table}[t]
	\centering
	\resizebox{1\columnwidth}{!}{%
	\begin{tabular}{cccccc}
		\toprule
         \multirow{2}{*}{Parser}&\multicolumn{2}{c}{Performance}&\multicolumn{3}{c}{Dataset}\\
                     \cmidrule(r){2-3} \cmidrule(r){4-6}
            & UAS   & LAS  & Restaurant& Laptop&Twitter  \\

        \hline
        Stanford      & 94.10 &  91.49 & 0.8133&0.7539&0.7283 \\
        Biaffine      & \textbf{95.74}& \textbf{94.08} & \textbf{0.8330}&\textbf{0.7742}&\textbf{0.7557} \\
        \bottomrule
	\end{tabular}
	}
	\caption{Results of R-GAT based on two different parsers, where \texttt{UAS} and \texttt{LAS} are metrics to evaluate the parsers and higher scores mean better performance.}\vspace{-0.0cm}
	\label{tab:parser}
\end{table}

\begin{table}[t]
\small
	\centering
	\begin{tabular}{p{1cm}p{1.8cm}p{1.2cm}<{\centering}cp{0.7cm}<{\centering}}
		\toprule
        Tree&Method     & Restaurant& Laptop&Twitter  \\
        \hline
        \multirow{2}{*}{Ordinary}
        &GAT&78.21&73.04&71.67\\
         &R-GAT  &79.91 &72.72&71.76\\
         \midrule
         \multirow{3}{*}{Reshaped }
        &GAT  &78.57 &72.10&71.82\\
        &R-GAT   &83.30&77.42&75.57\\
        &R-GAT$_{-n:con}$ &81.16&73.66&70.95\\
        \bottomrule
	\end{tabular}
    \caption{Results of ablation study, where ``Ordinary'' means using ordinary dependency trees, ``Reshaped'' denotes using the aspect-oriented trees, and ``*-n:con'' denote the aspect-oriented tree without using \texttt{n:con}.}\vspace{-0.4cm}
    \label{tab:ablation}
\end{table}

\begin{table*}[h]
\small
\centering
\begin{tabular}{p{2.2cm}p{0.5cm}p{12cm}}
\toprule
Category&(\%)&Example\\
\midrule
Neutral&46&No green beans, no egg, no anchovy dressing, no [nicoise olives]$_{neu}$, no red onion.\\
Comprehension&32&It took about 2 1/2 hours to be served our 2 [courses]$_{neg}$.\\
Advice&6& Try the [rose roll]$_{pos}$ (not on menu).\\
Double negation&6&But [dinner]$_{pos}$ here is never disappointing, even if the prices are a bit over the top.\\
\midrule
\end{tabular}
\begin{tabular}{p{2.2cm}p{0.5cm}p{12cm}}
\midrule
Neutral&50&Entrees include classics like lasagna, [fettuccine alfredo]$_{neu}$ and chicken parmigiana.\\
Comprehension&31&We requested they re-slice the [sushi]$_{pos}$, and it was returned to us in small cheese-like cubes.\\
Advice&5&Gave a [mojito]$_{pos}$ and sit in the back patio.\\
Double negation&3&And these are not small, wimpy fast food type [burgers]$_{pos}$ - these are real, full sized patties\\
\bottomrule
\end{tabular}
\caption{Results of error analysis of R-GAT and R-GAT+BERT on 100 misclassified examples from the Restaurant dataset. The reasons are classified into four categories, for which a sample is given. The upper table corresponds to the results of R-GAT and the lower one corresponds to R-GAT+BERT.}
\label{tab:badcase}
\end{table*}

\subsubsection{Ablation Study}
We further conduct an ablation study to evaluate the influence of the aspect-oriented dependency tree structure and the relational heads. We present the results on ordinary dependency trees for comparison. From table \ref{tab:ablation}, we can observe that R-GAT is improved by using the new tree structure on all three datasets, while GAT is only improved on the Restaurant and Twitter datasets. Furthermore, after removing the virtual relation \texttt{n:con}, the performance of R-GAT drops considerably. We manually examined the misclassified samples and found that most of them can be attributed to poor parsing results where aspects and their opinion words are incorrectly connected. This study validates that adding the \texttt{n:con} relation can effectively alleviate the parsing problem and allows our model to be robust. In this paper, the maximal number of \texttt{n} is set to 4 according to empirical tests. Other values of \texttt{n} are also explored but the results are not any better. This may suggest that words with too long dependency distances from the target aspect are unlikely to be useful for this task.

\subsubsection{Error Analysis}

To analyze the limitations of current ABSA models including ours, we randomly select 100 misclassified examples by two models (R-GAT and R-GAT+BERT) from the Restaurant dataset. After looking into these bad cases, we find the reasons behind can be classified into four categories. As shown in Table \ref{tab:badcase}, the primary reason is due to the misleading neutral reviews, most of which include an opinion modifier (words) towards the target aspect with a direct dependency connection. The second category is due to the difficulty in comprehension, which may demand deep language understanding techniques such as natural language inference. The third category is caused by the advice which only recommend or disrecommend people to try, with no obvious clues in the sentences indicating the sentiments. The fourth category is caused by double negation expression, which is also difficult for current models. Through the error analysis, we can note that although current models have achieved appealing progress, there are still some complicated sentences beyond their capabilities. There ought to be more advanced natural language processing techniques and learning algorithms developed to further address them.

\section{Conclusion}
In this paper, we have proposed an effective approach to encoding comprehensive syntax information for aspect-based sentiment analysis. We first defined a novel aspect-oriented dependency tree structure by reshaping and pruning an ordinary dependency parse tree to root it at a target aspect. We then demonstrated how to encode the new dependency trees with our relational graph attention network (R-GAT) for sentiment classification. Experimental results on three public datasets showed that the connections between aspects and opinion words can be better established with R-GAT, and the performance of GAT and BERT are significantly improved as a result. We also conducted an ablation study to validate the role of the new tree structure and the relational heads. Finally, an error analysis was performed on incorrectly-predicted examples, leading to some insights into this task.

\section*{Acknowledgments}

The work was supported by the Fundamental Research Funds for the Central Universities (No.19lgpy220) and the Program for Guangdong Introducing Innovative and Entrepreneurial Teams (No.2017ZT07X355). Part of this work was done when the first author was an intern at Alibaba. 

\bibliography{acl2020}
\bibliographystyle{acl_natbib}

\end{document}